\documentclass[conference]{IEEEtran}
\IEEEoverridecommandlockouts
\usepackage{cite}
\usepackage{amsmath,amssymb,amsfonts}
\usepackage{algorithmic}
\usepackage{graphicx}
\usepackage{textcomp}
\usepackage{xcolor}
\usepackage{caption}
\usepackage{algorithm}
\usepackage{algorithmic}

\usepackage{booktabs}

\def\BibTeX{{\rm B\kern-.05em{\sc i\kern-.025em b}\kern-.08em
    T\kern-.1667em\lower.7ex\hbox{E}\kern-.125emX}}
\begin{document}

\title{Neural Network Interpretability With Layer-Wise Relevance Propagation: Novel Techniques for Neuron Selection and Visualization
\\

}

\author{
  \IEEEauthorblockN{
    Deepshikha Bhati\textsuperscript{1}, 
    Fnu Neha\textsuperscript{1}, 
    Md Amiruzzaman\textsuperscript{2}, 
    Angela Guercio\textsuperscript{1}, 
    Deepak Kumar Shukla\textsuperscript{3},
    Ben Ward\textsuperscript{1}, 
  }
  \IEEEauthorblockA{
    \textsuperscript{1}Kent State University, Kent, Ohio, USA \\
    \{dbhati, neha, aguercio, bward29\}@kent.edu \\
    \textsuperscript{2}West Chester University,  West Chester, PA, USA \\
    mamiruzzaman@wcupa.edu\\
     \textsuperscript{3}Rutgers University, Newark, NJ, USA \\
    ds1640@scarletmail.rutgers.edu
  }
}

\maketitle

\begin{abstract}
Interpreting complex neural networks is crucial for understanding their decision-making processes, particularly in applications where transparency and accountability are essential. This proposed method addresses this need by focusing on Layer-wise Relevance Propagation (LRP), a technique used in explainable artificial intelligence (XAI) to attribute neural network outputs to input features through backpropagated relevance scores. Existing LRP methods often struggle with precision in evaluating individual neuron contributions. To overcome this limitation, we present a novel approach that improves the parsing of selected neurons during LRP backward propagation, using the Visual Geometry Group 16 (VGG16) architecture as a case study. Our method creates neural network graphs to highlight critical paths and visualizes these paths with heatmaps, optimizing neuron selection through accuracy metrics like Mean Squared Error (MSE) and Symmetric Mean Absolute Percentage Error (SMAPE). Additionally, we utilize a deconvolutional visualization technique to reconstruct feature maps, offering a comprehensive view of the network’s inner workings. Extensive experiments demonstrate that our approach enhances interpretability and supports the development of more transparent artificial intelligence (AI) systems for computer vision applications. This advancement has the potential to improve the trustworthiness of AI models in real-world machine vision applications, thereby increasing their reliability and effectiveness.

\end{abstract}


\section{Introduction}
Understanding the decision-making processes of complex neural networks (NN) has become increasingly critical as these models are deployed in real-world applications such as autonomous driving, medical diagnostics, and financial forecasting. Interpretability is particularly crucial in these domains to ensure transparency, build trust, and facilitate regulatory compliance. Layer-wise Relevance Propagation (LRP) is a prominent technique aimed at interpreting the decisions of NN by propagating the output backward through the layers to the input \cite{bach2015pixel}. This technique highlights which pixels or intermediate neurons significantly contribute to the final decision by generating \textit{relevance} values (i.e., $R$), for each pixel or neuron, thereby enabling a detailed examination of the network's inner working process.

Despite its effectiveness, LRP's practical application and visualization present significant challenges. Traditional methods do not provide clear and scalable interpretability, especially when dealing with large-scale NNs and high-dimensional data. To address these limitations, we propose a novel approach that leverages LRP's backward propagation capabilities to evaluate and visualize interpretability of convolutional neural networks (CNNs). Our approach enhances interpretability by identifying and highlighting the most relevant neurons and their paths through network layers.

Our main contributions to this paper include:
\begin{enumerate}
\item Generating NN graphs to identify significant paths (connecting activated neurons layer by layer);
\item Developing algorithms for optimized path selection (the path follows the activation from the start to end the end);
\item Creating visualization heatmaps for the selected $k$-paths (out of total $n$-number of paths to selected $k$-paths);

\end{enumerate}
In this paper, we detail the fundamentals of LRP, the specific steps involved in its computation, and the application of our proposed methods to enhance interpretability in NN. We also introduce accuracy metrics such as Mean Squared Error (MSE) and Symmetric Mean Absolute Percentage Error (SMAPE) to evaluate our approach, along with a comprehensive algorithm for optimized path selection in neuron visualization.

The paper is organized as follows: Section 2 reviews related work on NN interpretability and LRP. Section 3 covers LRP, and the Visual Geometry Group 16 (VGG16) architecture. Section 4 outlines our proposed approach. Section 5 presents a case study based on VGG16. Finally, section 6 concludes with a summary of findings and future directions.

\section{Related Work}

Interpreting NNs, especially CNNs, is a critical research area due to their complexity and extensive use. CNNs excel in image classification and segmentation tasks \cite{nazir2023survey}, and explainable AI (XAI) techniques in biomedical imaging \cite{tjoa2020survey, huang2020survey, bhati2024survey}, however, their internal representations and decision-making remain challenging to understand \cite{zeiler2014visualizing}. Techniques like occlusion and optimized activation excitement have been developed for model visualization and explanation \cite{iwana2019explaining}.

LRP assigns relevance scores to input features based on their contribution to the network's output \cite{bach2015pixel}, and has been applied in image classification \cite{samek2019towards}, medical imaging \cite{binder2016layer}, and fine-grained visual tasks. Montavon et al. enhanced LRP with rules to improve interpretability \cite{montavon2017explaining}, though scaling LRP to larger models remains a challenge. Lapuschkin et al. proposed techniques to visualize heatmaps in large models, underscoring the need for scalable methods \cite{lapuschkin2019unmasking}.

In fine-grained classification, LRP has been used for analyzing detailed image features. Arquilla et al. applied LRP to bird species classification, highlighting fine-grained features like feather patterns \cite{arquilla2024exploring}.
Other interpretability methods include Grad-CAM, which highlights important regions using gradients \cite{selvaraju2017grad}; LIME, providing model-agnostic explanations through interpretable surrogate models \cite{ribeiro2016should}; and SHAP, which offers consistent explanations based on cooperative game theory \cite{lundberg2017unified}. Despite these advancements, scalable interpretability for large NNs remains difficult due to their complexity.

\begin{figure}
    \centering
    \includegraphics[width=1\linewidth]{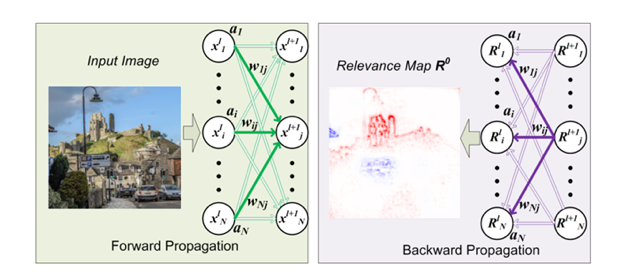}
    \caption{LRP computational process with two phases: forward propagation of activation and backward propagation of relevance.}
    \label{fig:lrp}
\end{figure}

Our work extends LRP by introducing new techniques for generating NN graphs, visualizing heatmaps, and optimizing relevance scores, aiming to provide a more comprehensive and scalable solution for NN interpretability, addressing gaps left by existing methods.

\section{Method and Model Overview}
In this section, we present an overview of the methodologies and models central to our research. We start with a discussion on Layer-wise Relevance Propagation (LRP), which is important for understanding NN decision-making processes. Next, we describe the VGG16 architecture, outlining its design principles and its application in image classification.

\begin{figure*}
    \centering
    \includegraphics[width=1\linewidth]{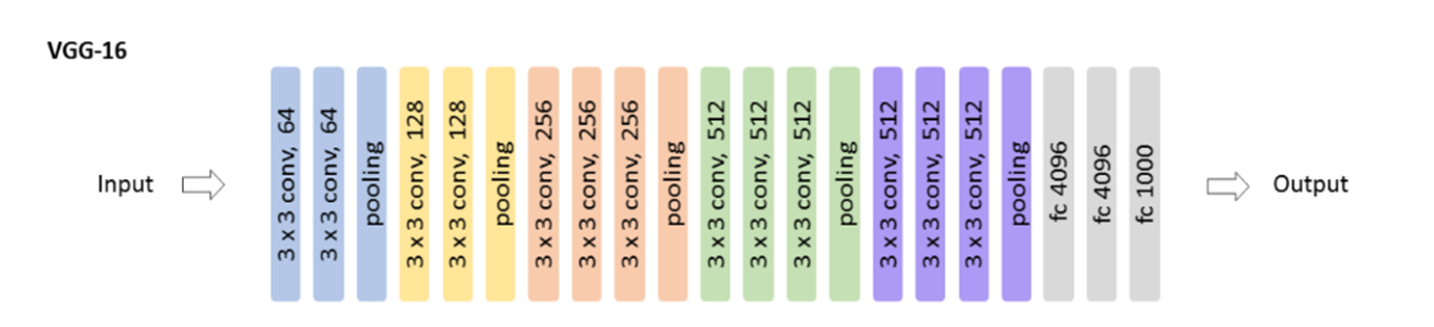}
    \caption{Architecture of VGG16.}
    \label{fig:vgg16}
\end{figure*}

\subsection{Layer-wise Relevance Propagation (LRP) Fundamentals}
As mentioned earlier, LRP provides insights into NN decision-making by tracing the network's weights and activations from the forward pass and propagating the output relevance backward. This process ensures that the total relevance at the output layer equals the sum of the relevance (i.e., $R$) values at the input layer, maintaining the integrity of relevance scores throughout the network. Figure \ref{fig:lrp} illustrates this conservation property.

For a network output vector \( y \) of size \( M \), corresponding to \( M \) classes, we select one class \( c \) to explain. The relevance of the output neuron for class \( c \) is set equal to its activation, while the relevance of all other output neurons is zero. The relevance scores \( R_j \) for neurons \( j \) in layer \( k \) are computed using the following basic LRP rule:

\[
R_j = \sum_k \frac{a_j w_{jk}}{\sum_{j} a_j w_{jk}} R_k
\]

where \( a \) denotes neuron activations and \( w \) represents the weights between neurons in consecutive layers. The numerator \( a_j w_{jk} \) measures neuron \( j \)’s influence on neuron \( k \), and the denominator normalizes these contributions to ensure relevance conservation.

The LRP computation involves four main steps:

\begin{itemize}
    \item \textbf{Forward Pass:}
    \[
    z_k = \epsilon + \sum_{j} a_j \rho(w_{jk})
    \]
    This step aggregates influences for each higher-layer neuron. For ReLU layers, this step mirrors the regular forward pass.
    
    \item \textbf{Element-wise Division:}
    \[
    s_k = \frac{R_k}{z_k}
    \]
    Relevance in the higher layer is divided by its total influence \( z_k \) to maintain conservation.
    
    \item \textbf{Backward Pass:}
    \[
    C_j = \sum_k \rho(w_{jk}) s_k
    \]
    Computes a quantity \( C_j \) for every neuron in the preceding layer, reflecting the relevance passed down from the higher layer.
    
    \item \textbf{Element-wise Product:}
    \[
    R_j = a_j * C_j
    \]
    Relevance is multiplied by neuron activation to determine the neuron's relevance score.
\end{itemize}

Different propagation rules are applied based on the layer being analyzed:

\begin{itemize}
    \item \textbf{Input Layer:} The Deep Taylor Decomposition rule provides accurate relevance mapping using both the lowest and highest admissible pixel values:
    \[
    R_i = \frac{(x_i w_{ij} - l_i w_{ij}^+ - h_i w_{ij}^-)}{\sum_i (x_i w_{ij} - l_i w_{ij}^+ - h_i w_{ij}^-)} R_j
    \]
    
    \item \textbf{Lower Layers:} The LRP-\(\gamma\) rule generates smooth relevance maps with minimal noise, emphasizing positive contributions:
    \[
    R_j = \sum_k \frac{a_j w_{jk}^+}{\sum_{j} a_j w_{jk}^+} R_k
    \]
    
    \item \textbf{Higher Layers:} The LRP-\(\epsilon\) rule stabilizes relevance scores by adding a small constant \(\epsilon\):
    \[
    R_j = \sum_k \frac{a_j w_{jk}}{\epsilon + \sum_{j} a_j w_{jk}} R_k
    \]
    
    \item \textbf{Output Layer:} The LRP-0 rule propagates relevance directly without modification.
\end{itemize}

\begin{figure*}[h]
    \centering
    \includegraphics[width=0.7\linewidth]{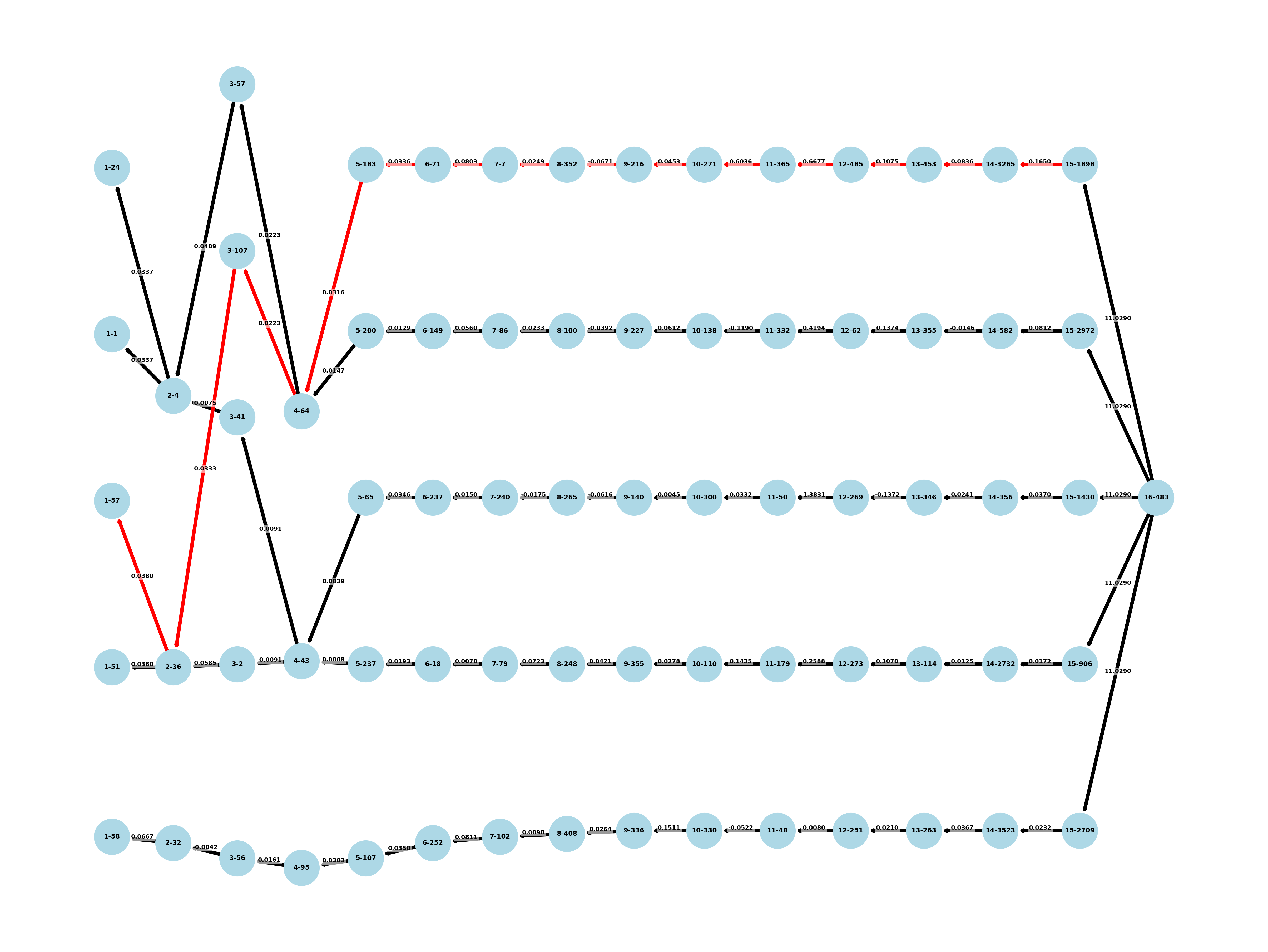}
    \caption{Highlight longest paths in red, emphasizing their relevance in backward propagation (LRP relevance scores).}
    \label{fig:graph}
\end{figure*}

\begin{figure*}[h]
    \centering
    \includegraphics[width=0.7\linewidth]{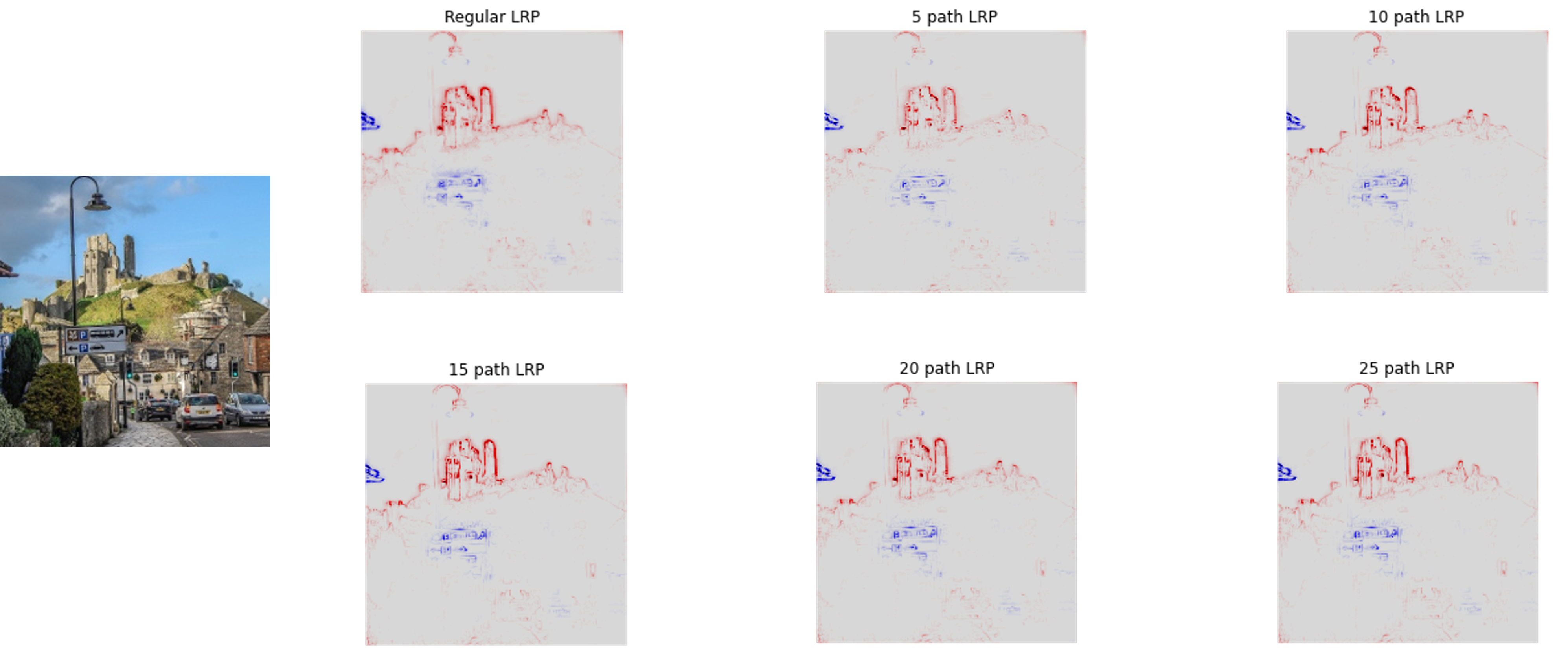}
    \caption{Visualization heatmap based on k-path selection.}
    \label{fig:heatmap_path}
\end{figure*}

\subsection{Visual Geometry Group 16 (VGG16) Architecture}
The VGG16 is a CNN for image classification with a straightforward yet effective design, consisting of 16 layers: 13 convolutional and 3 fully connected (FC) layers (see Figure \ref{fig:vgg16}). Convolutional layers use 3x3 kernels with a stride of 1, followed by max-pooling layers to downsample feature maps. The flattened output is processed through three FC layers with 4096, 4096, and 1000 channels, respectively. VGG16 has approximately 138 million parameters and achieves 92.70\% top-5 accuracy on the ILSVRC2014 dataset \cite{liu2015very}-\cite{russakovsky2015imagenet}.

\section{Proposed Approach}

NNs process input data through multiple layers, each layer extracting specific features. To improve interpretability, we propose an approach that focuses on analyzing the relevance of detected features within these hidden layers by leveraging LRP and visualizing the most significant paths.

\subsection{Neural Network Graph Generation}
We first generate a NN graph to identify and highlight important paths within the network, as shown in Figure 3, by selecting the optimal path using Algorithm~\ref{alg:get_optimizer}.
\begin{itemize}
    \item \textbf{Path Identification}: The \textit{longest paths} in the graph are identified, which correspond to the most critical connections based on LRP relevance scores.
    \item \textbf{Path Visualization}: These longest paths are highlighted in \textit{red}, emphasizing their importance in backward propagation.
\end{itemize}

\subsection{Visualization Heatmap}

To further interpret the network's decisions, heatmaps are generated based on the selected paths.

\begin{itemize}
    \item \textbf{Heatmap Generation}: Heatmaps are created to correspond to the selected paths, providing a visual representation of the relevance scores as shown in Figure \ref{fig:heatmap_path}.
    \item \textbf{Performance Metrics}: To assess the effectiveness of our approach, we utilize performance metrics such as MSE and SMAPE. We calculate these metrics to evaluate the accuracy of predictions and optimize the number of paths needed for prediction.

\begin{itemize}
    \item \textbf{MSE}: It measures the average squared difference between predicted and actual values.
    \[
    \text{MSE} = \frac{1}{n} \sum_{i=1}^n (D_{\text{pre}} - D_{\text{act}})^2
    \]

    \item \textbf{SMAPE}: It provides a percentage-based accuracy measurement, offering a unit-free comparison of prediction errors.
    \[
    \text{SMAPE} = \frac{100\%}{n} \sum_{i=1}^n \frac{|D_{\text{pre}} - D_{\text{act}}|}{(|D_{\text{pre}}| + |D_{\text{act}}|)/2}
    \]
\end{itemize}
\end{itemize}

\subsection{Optimized Path Selection Algorithm}

Our approach includes an optimized path selection algorithm designed to identify and visualize the most contributing neurons, as detailed in Algorithm~\ref{alg:get_optimizer}:

\algsetup{linenosize=\small}
\begin{algorithm}[H]
\caption{GetOptimizer}
\label{alg:get_optimizer}
\begin{algorithmic}[1]
    \STATE \textbf{Input:} $forwardResult$, $backwardResult$, $index$
    \STATE \textbf{Output:} $result$
    \STATE $p \gets dim2$
    \STATE $neutralValue \gets 0.0$
    \STATE $diffResult \gets forwardResult[index] - backwardResult[index]$
    \STATE $sumResult \gets \sum_{axis=p} diffResult$
    \STATE $meanValue \gets \text{mean}(sumResult)$
    \STATE $stdValue \gets \text{std}(sumResult)$
    \STATE $result \gets \text{where}(sumResult > (meanValue - stdValue), neutralValue, sumResult)$
    \RETURN $result$
\end{algorithmic}
\end{algorithm}

\textbf{Explanation:} Algorithm~\ref{alg:get_optimizer}, \textit{GetOptimizer}, is central to our approach for optimizing the path selection in NNs. The algorithm begins by taking the forward and backward pass results of a NN and an index as inputs. The primary goal is to compute a difference between these results, which serves as a measure of how each neuron contributes to the network's final prediction.

First, the algorithm calculates the difference between the forward and backward results at a specified index, highlighting neurons that impact the prediction. It then sums this difference across a dimension, $p$, to aggregate neuron contributions.
The mean and standard deviation of the summed results are computed to identify relevant neurons. Neurons with contributions above one standard deviation below the mean are retained, while others are set to a neutral value (zero).
The result, showing the optimized path of contributing neurons, highlights the most significant neurons in the network's decision-making process, enhancing model interpretability.

As shown in Figure~\ref{fig:graph}, this approach allows for a clear visualization of the NN's decision-making process by focusing on the most relevant neurons. The figure shows how the algorithm reduces the complexity of the network's decision path by filtering out less significant contributions, thus providing a more interpretable summary of the model's reasoning.

\begin{figure*}[t]
    \centering
    \includegraphics[width=0.75\textwidth]{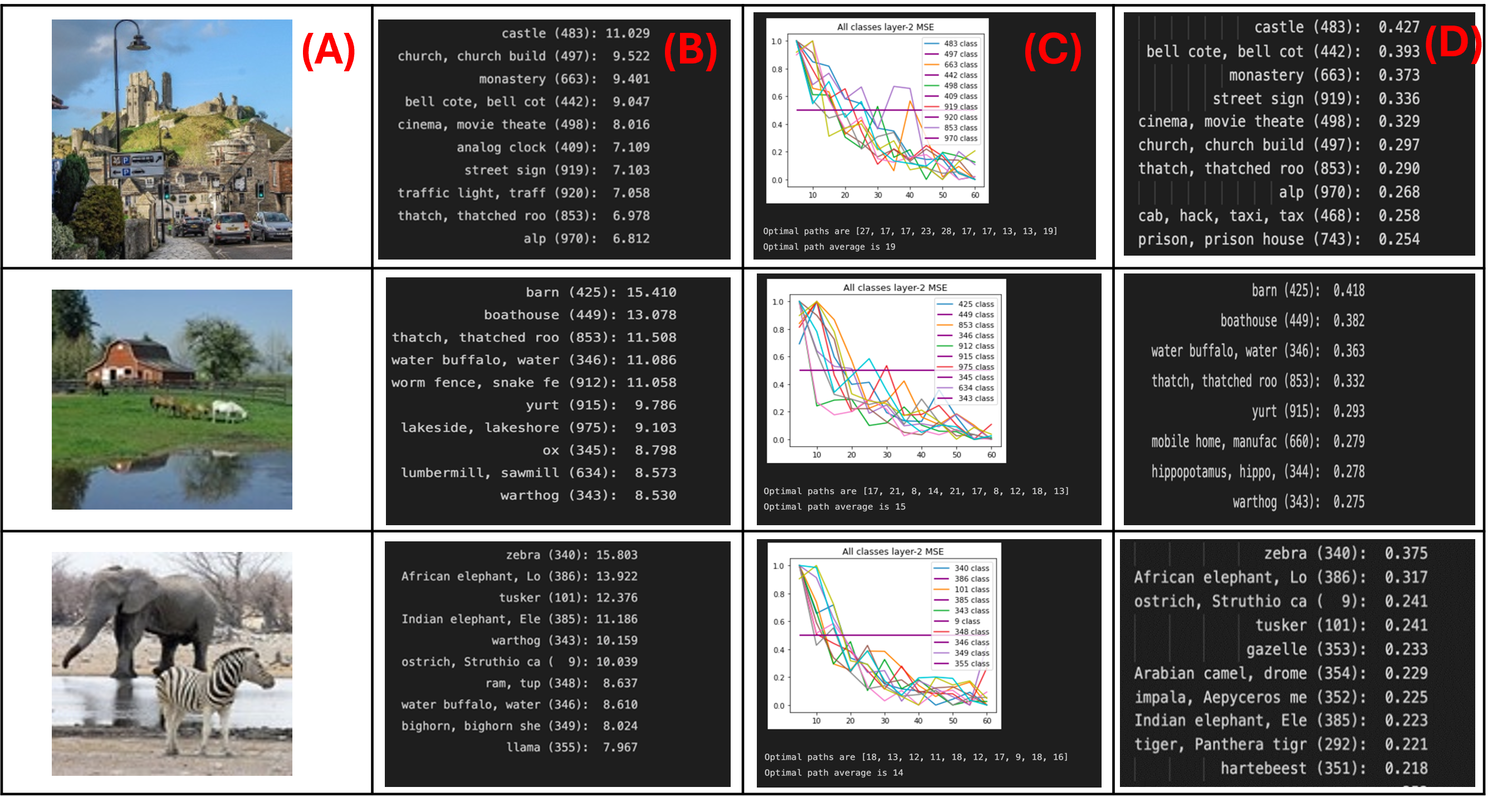}  
    \vspace{2mm}  
    \caption{(A) Visualization of Original Images, (B) Model Predictions, (C) Graphs, and (D) Back-predictions for Comparative Analysis in Three Categories: Castle, Barn, and Zebra.}
    \label{fig:analysis}
    \vspace{2mm}  
\end{figure*}

\begin{figure*}
    \centering
    \includegraphics[width=0.75\linewidth]{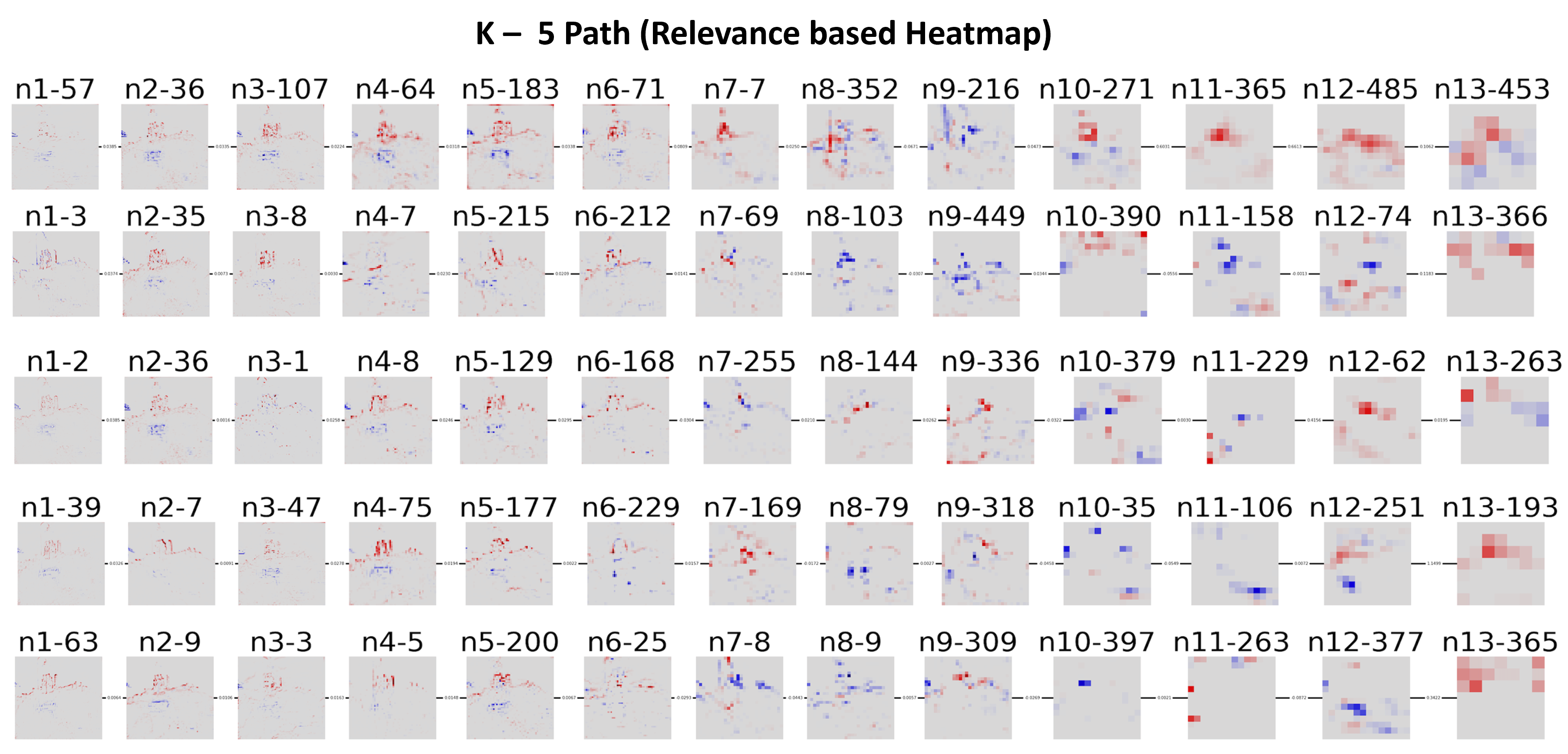}
    \caption{K –  5 Path (Relevance based Heatmap)}
    \label{fig:K5LRP}
\end{figure*}

\begin{figure*}
    \centering
    \includegraphics[width=0.75\linewidth]{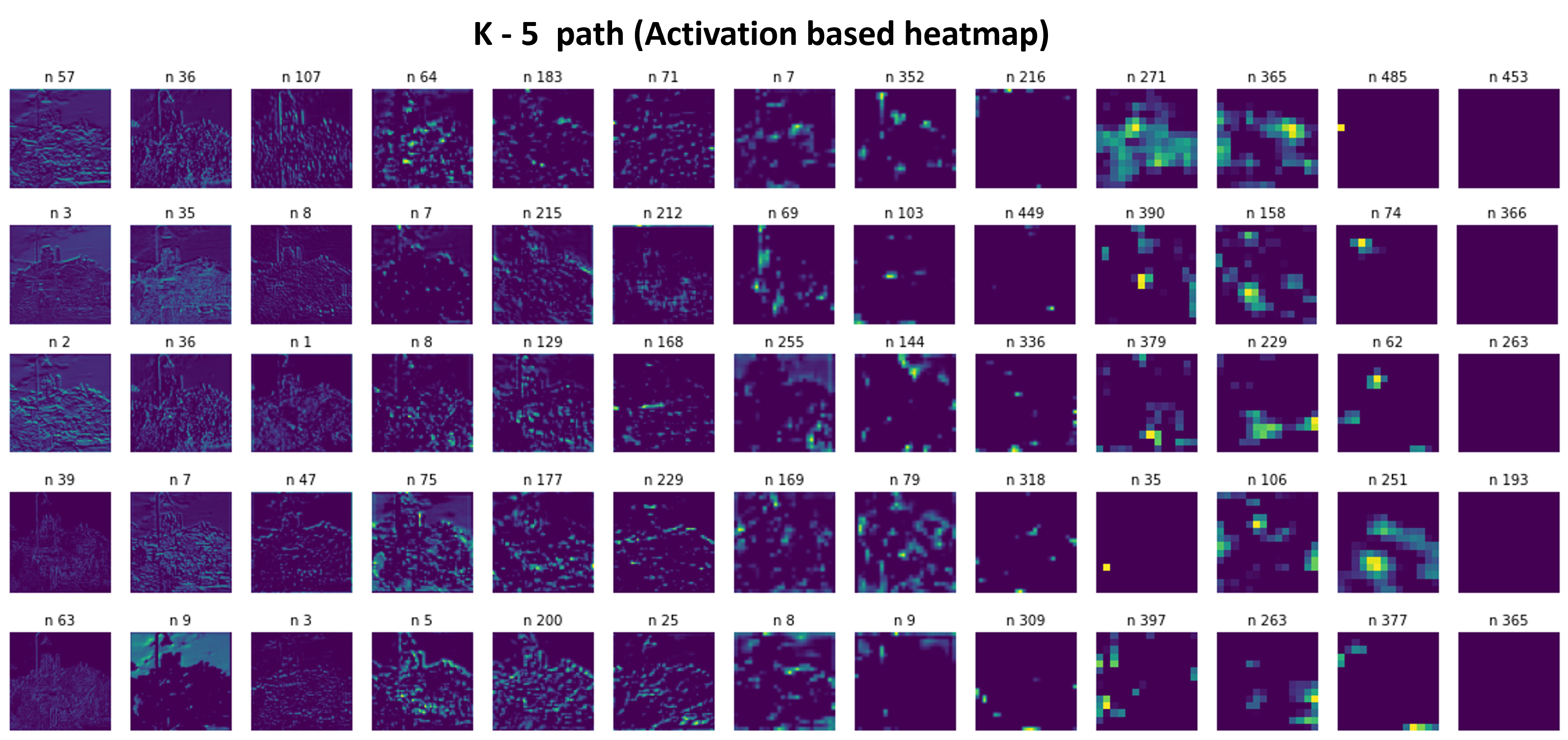}
    \caption{K –  5 Path (Activation based heatmap)}
    \label{fig:K5Activation}
\end{figure*}

\subsection{Optimizing Threshold and Visualization Techniques}
In this subsection, we present our approach to determining the optimal threshold for path selection, which involves averaging MSE values to identify the most contributing neurons for predictions. Additionally, we utilize deconvolutional methods to reconstruct feature maps from input images, allowing us to visualize significant neurons through backward operations. This process is detailed in Algorithm~\ref{alg:feature_map_reconstruction} and Algorithm~\ref{alg:forward_operation}.

\begin{enumerate}
    \item \textbf{Deconvolution Process}
    \begin{itemize}
        \item Set all activations to zero except the selected neuron.
        \item Apply unpooling and deconvolution to reconstruct features in image space.
    \end{itemize}

    \item \textbf{Feature Map Reconstruction}
    \begin{itemize}
        \item Create sub-networks for convolution and deconvolution layers.
        \item Initialize deconvolution layers with the weights from corresponding convolution layers.
        \item Store max locations for unpooling.
    \end{itemize}
\end{enumerate} 

\setlength{\algorithmicindent}{0.5em}
\algsetup{linenosize=\small}
\begin{algorithm}[H]
\caption{Feature Map Reconstruction}
\label{alg:feature_map_reconstruction}
\begin{algorithmic}[1]
    \STATE \textbf{Input:} Image $img$, Layer $layer\_idx$, Neuron $path\_nrn$
    \STATE \textbf{Output:} Reconstructed image
    \STATE $pooled\_out, max\_idx \gets \text{MaxPool}(layer\_input)$
    \STATE Zero neurons in $pooled\_out$ except $path\_nrn$
    \STATE $unpooled\_out \gets \text{MaxUnpool}(pooled\_out, max\_idx)$
    \STATE $deconv\_out \gets \text{ConvTrans}(unpooled\_out)$
    \RETURN $deconv\_out$
\end{algorithmic}
\end{algorithm}

\textbf{Explanation:} Algorithm~\ref{alg:feature_map_reconstruction}, \textit{Feature Map Reconstruction}, reconstructs images by reversing pooling and convolution operations, allowing for the visualization of specific neuron contributions to the classification process. This method enables a detailed understanding of how individual neurons influence the final prediction by tracing back the activations through the network layers.

\algsetup{linenosize=\small}
\begin{algorithm}[H]
\caption{Forward Operation of the Network}
\label{alg:forward_operation}
\begin{algorithmic}[1]
    \STATE \textbf{Input:} Input data $x$, Layer index $layer\_idx$, Path neuron index $path\_neuron$
    \STATE \textbf{Output:} Reconstructed feature maps

    \FOR{$idx \gets 0$ \TO $layer\_idx$}
        \IF{$conv\_sub\_network[idx]$ \text{ is } $MaxPool2d$}
            \STATE $x, loc \gets conv\_sub\_network[idx](x)$
            \STATE \text{Store } $loc$ \text{ in } $switches[idx]$
        \ELSE
            \STATE $x \gets conv\_sub\_network[idx](x)$
        \ENDIF
    \ENDFOR

    \STATE \text{Set feature maps in } $x$ \text{ to zero except for } $path\_neuron$

    \FOR{$idx \gets (30 - layer\_idx)$ \TO $length(deconv\_sub\_network)$}
        \IF{$deconv\_sub\_network[idx]$ \text{ is } $MaxUnpool2d$}
            \STATE $x \gets deconv\_sub\_network[idx](x, switches[30 - idx])$
        \ELSE
            \STATE $x \gets deconv\_sub\_network[idx](x)$
        \ENDIF
    \ENDFOR

    \RETURN $x$
\end{algorithmic}
\end{algorithm}

\textbf{Explanation:} Algorithm~\ref{alg:forward_operation}, \textit{Forward Operation}, processes input data through the network to a specified layer, focusing on selected neurons. It then reconstructs feature maps by tracing activations backward, providing a clear visualization of the most influential neurons and their contributions to the final decision. This approach effectively interprets neural network predictions by highlighting and visualizing key neurons through optimized path selection, aiding in understanding complex neural networks.

Figure~\ref{fig:analysis} demonstrates the effectiveness of our methods, showing how the optimized threshold and visualization techniques allow for clear identification of significant neurons within the network. The figure illustrates the reconstructed feature maps and the paths that contribute most to the network's decision-making process, thereby enhancing the interpretability of the model's predictions.

\section{EXPERIMENTS}
\subsection{Datasets}
For our experiments, we used the ImageNet dataset \cite{deng2012imagenet}, a widely recognized benchmark in image classification. ImageNet contains over 14 million labeled images spanning over 20,000 categories, offering a diverse and extensive resource for training and evaluating deep learning models. We experiment on the VGG16 model, pre-trained on ImageNet, to leverage its learned features for our interpretability analysis.

\subsection{Case Study}
To illustrate our approach, we consider an example where John, a researcher, analyzes a castle image from the ImageNet dataset using our method.

\begin{itemize}
    \item \textbf{Initial Exploration:} 
    First, John examines the VGG16 model's predictions on the castle image, focusing on understanding which image regions most influence the model's decision.

    \item \textbf{Relevance-Based Analysis:}  
He applies LRP to the VGG16 model to derive relevance scores for each neuron. Figure~\ref{fig:K5LRP} presents the K-5 Path heatmap, highlighting areas of the image that significantly impact the model's prediction. These relevance scores show which parts of the castle—such as the towers, walls, and windows—are crucial for the model’s decision-making.

\item \textbf{Activation-Based Analysis:}  
Next, John analyzes neuron activations in VGG16. Figure~\ref{fig:K5Activation} displays the K-5 Path heatmap based on activations, showing how specific neurons respond to the castle image. This helps him understand the broader regions the network is focused on, including features like the overall shape of the castle and its structural elements.

\item \textbf{Comparative Analysis:}  
In Figure~\ref{fig:analysis}, John compares the original image, model predictions, and back-prediction graphs. He observes how the relevance and activation heatmaps align with the model's output. The comparative analysis allows him to see the overlap between the regions of high relevance and strong activation, providing insights into how VGG16 interprets and processes the castle image.

\end{itemize}

Through this analysis, John discovers that our method provides a comprehensive understanding of neural network predictions. The combination of relevance and activation-based analyses offers a nuanced view of the decision-making process, emphasizing both the significance of specific neurons and their activation patterns. This dual approach helps clarify not only where the network is focusing but also why certain features, such as the architectural elements of the castle, are crucial for classification.

\section{Conclusion and Future Work}
This paper explored LRP applied to the VGG16 architecture, focusing on feature contributions by analyzing neuron selections during LRP's backward propagation. We developed methods to generate NN graphs, visualize heatmaps, and reconstruct feature maps via deconvolution. Our findings show that precise neuron selection enhances NN interpretability, offering clearer insights into model processing.

Future work will extend this approach to architectures such as residual networks and transformer-based models to test its generalizability. Integrating LRP with other techniques such as SHAP or Grad-CAM could provide complementary insights. Automating neuron selection through optimization strategies and conducting user studies to evaluate visualization impact are also planned. Investigating the relationship between interpretability, model robustness, and fairness is a key area for advancing transparent and trustworthy AI systems.

\bibliographystyle{IEEEtran}
\bibliography{main.bib}

\begin{thebibliography}{10}
\providecommand{\url}[1]{#1}
\csname url@samestyle\endcsname
\providecommand{\newblock}{\relax}
\providecommand{\bibinfo}[2]{#2}
\providecommand{\BIBentrySTDinterwordspacing}{\spaceskip=0pt\relax}
\providecommand{\BIBentryALTinterwordstretchfactor}{4}
\providecommand{\BIBentryALTinterwordspacing}{\spaceskip=\fontdimen2\font plus
\BIBentryALTinterwordstretchfactor\fontdimen3\font minus \fontdimen4\font\relax}
\providecommand{\BIBforeignlanguage}[2]{{%
\expandafter\ifx\csname l@#1\endcsname\relax
\typeout{** WARNING: IEEEtran.bst: No hyphenation pattern has been}%
\typeout{** loaded for the language `#1'. Using the pattern for}%
\typeout{** the default language instead.}%
\else
\language=\csname l@#1\endcsname
\fi
#2}}
\providecommand{\BIBdecl}{\relax}
\BIBdecl

\bibitem{bach2015pixel}
S.~Bach, A.~Binder, G.~Montavon, F.~Klauschen, K.-R. M{\"u}ller, and W.~Samek, ``On pixel-wise explanations for non-linear classifier decisions by layer-wise relevance propagation,'' \emph{PloS one}, vol.~10, no.~7, p. e0130140, 2015.

\bibitem{nazir2023survey}
S.~Nazir, D.~M. Dickson, and M.~U. Akram, ``Survey of explainable artificial intelligence techniques for biomedical imaging with deep neural networks,'' \emph{Computers in Biology and Medicine}, vol. 156, p. 106668, 2023.

\bibitem{tjoa2020survey}
E.~Tjoa and C.~Guan, ``A survey on explainable artificial intelligence (xai): Toward medical xai,'' \emph{IEEE transactions on neural networks and learning systems}, vol.~32, no.~11, pp. 4793--4813, 2020.

\bibitem{huang2020survey}
X.~Huang, D.~Kroening, W.~Ruan, J.~Sharp, Y.~Sun, E.~Thamo, M.~Wu, and X.~Yi, ``A survey of safety and trustworthiness of deep neural networks: Verification, testing, adversarial attack and defence, and interpretability,'' \emph{Computer Science Review}, vol.~37, p. 100270, 2020.

\bibitem{bhati2024survey}
\BIBentryALTinterwordspacing
D.~Bhati, F.~Neha, and M.~Amiruzzaman, ``A survey on explainable artificial intelligence (xai) techniques for visualizing deep learning models in medical imaging,'' \emph{Preprints}, 2024, 2024080765. [Online]. Available: \url{https://doi.org/10.20944/preprints202408.0765.v1}
\BIBentrySTDinterwordspacing

\bibitem{zeiler2014visualizing}
M.~D. Zeiler and R.~Fergus, ``Visualizing and understanding convolutional networks,'' in \emph{Computer Vision--ECCV 2014: 13th European Conference, Zurich, Switzerland, September 6-12, 2014, Proceedings, Part I 13}.\hskip 1em plus 0.5em minus 0.4em\relax Springer, 2014, pp. 818--833.

\bibitem{iwana2019explaining}
B.~K. Iwana, R.~Kuroki, and S.~Uchida, ``Explaining convolutional neural networks using softmax gradient layer-wise relevance propagation,'' in \emph{2019 IEEE/CVF International Conference on Computer Vision Workshop (ICCVW)}.\hskip 1em plus 0.5em minus 0.4em\relax IEEE, 2019, pp. 4176--4185.

\bibitem{samek2019towards}
W.~Samek and K.-R. M{\"u}ller, ``Towards explainable artificial intelligence,'' \emph{Explainable AI: interpreting, explaining and visualizing deep learning}, pp. 5--22, 2019.

\bibitem{binder2016layer}
A.~Binder, G.~Montavon, S.~Lapuschkin, K.-R. M{\"u}ller, and W.~Samek, ``Layer-wise relevance propagation for neural networks with local renormalization layers,'' in \emph{Artificial Neural Networks and Machine Learning--ICANN 2016: 25th International Conference on Artificial Neural Networks, Barcelona, Spain, September 6-9, 2016, Proceedings, Part II 25}.\hskip 1em plus 0.5em minus 0.4em\relax Springer, 2016, pp. 63--71.

\bibitem{montavon2017explaining}
G.~Montavon, S.~Lapuschkin, A.~Binder, W.~Samek, and K.-R. M{\"u}ller, ``Explaining nonlinear classification decisions with deep taylor decomposition,'' \emph{Pattern recognition}, vol.~65, pp. 211--222, 2017.

\bibitem{lapuschkin2019unmasking}
S.~Lapuschkin, S.~W{\"a}ldchen, A.~Binder, G.~Montavon, W.~Samek, and K.-R. M{\"u}ller, ``Unmasking clever hans predictors and assessing what machines really learn,'' \emph{Nature communications}, vol.~10, no.~1, p. 1096, 2019.

\bibitem{arquilla2024exploring}
K.~Arquilla, I.~D. Gajera, M.~Darling, D.~Bhati, A.~Singh, and A.~Guercio, ``Exploring fine-grained feature analysis for bird species classification using layer-wise relevance propagation,'' in \emph{2024 IEEE World AI IoT Congress (AIIoT)}.\hskip 1em plus 0.5em minus 0.4em\relax IEEE, 2024, pp. 625--631.

\bibitem{selvaraju2017grad}
R.~R. Selvaraju, M.~Cogswell, A.~Das, R.~Vedantam, D.~Parikh, and D.~Batra, ``Grad-cam: Visual explanations from deep networks via gradient-based localization,'' in \emph{Proceedings of the IEEE international conference on computer vision}, 2017, pp. 618--626.

\bibitem{ribeiro2016should}
M.~T. Ribeiro, S.~Singh, and C.~Guestrin, ``" why should i trust you?" explaining the predictions of any classifier,'' in \emph{Proceedings of the 22nd ACM SIGKDD international conference on knowledge discovery and data mining}, 2016, pp. 1135--1144.

\bibitem{lundberg2017unified}
S.~M. Lundberg and S.-I. Lee, ``A unified approach to interpreting model predictions,'' \emph{Advances in neural information processing systems}, vol.~30, 2017.

\bibitem{liu2015very}
S.~Liu and W.~Deng, ``Very deep convolutional neural network based image classification using small training sample size,'' in \emph{2015 3rd IAPR Asian conference on pattern recognition (ACPR)}.\hskip 1em plus 0.5em minus 0.4em\relax IEEE, 2015, pp. 730--734.

\bibitem{russakovsky2015imagenet}
O.~Russakovsky, J.~Deng, H.~Su, J.~Krause, S.~Satheesh, S.~Ma, Z.~Huang, A.~Karpathy, A.~Khosla, M.~Bernstein \emph{et~al.}, ``Imagenet large scale visual recognition challenge,'' \emph{International journal of computer vision}, vol. 115, pp. 211--252, 2015.

\bibitem{deng2012imagenet}
\BIBentryALTinterwordspacing
J.~Deng, A.~C. Berg, S.~Satheesh, H.~Su, A.~Khosla, and L.~Fei-Fei, ``The imagenet large scale visual recognition challenge 2012 (ilsvrc2012),'' 2012, accessed: 2015-04-01. [Online]. Available: \url{http://www.image-net.org/challenges/LSVRC/2012/}
\BIBentrySTDinterwordspacing

\end{thebibliography}

\end{document}